# A New Image Compression by Gradient Haar Wavelet

Yaser Sadra, *Department of computer sciences, Shandiz Institute of Higher Education, Mashhad, Iran*

*Abstract*—With the development of human communications the usage of Visual Communications has also increased. The advancement of image compression methods is one of the main reasons for the enhancement. This paper first presents main modes of image compression methods such as JPEG and JPEG2000 without mathematical details. Also, the paper describes gradient Haar wavelet transforms in order to construct a preliminary image compression algorithm. Then, a new image compression method is proposed based on the preliminary image compression algorithm that can improve standards of image compression. The new method is compared with original modes of JPEG and JPEG2000 (based on Haar wavelet) by image quality measures such as MAE, PSNAR, and SSIM. The image quality and statistical results confirm that can boost image compression standards. It is suggested that the new method is used in a part or all of an image compression standard.

## I. INTRODUCTION

NOWADAYS, digital images and videos play a significant role in human communications [1-6]. With the advancement of sciences such as social networks, telecommunication and deep space communication, the amount of data that needs to be transferred increases therefore it causes slow communication and expensive storage [7-10]. In field of the deep space communication, The National Aeronautics and Space Administration (NASA) say just for the Mars Reconnaissance Orbiter (MRO) had returned more than 298 terabits of data as of March 2016 [11]. Also, the NASA expresses with its maximum data rate of 5.2 megabits per second (Mbps), MRO requires 7.5 hours to empty its onboard recorder, and 1.5 hours to send a single HiRISE (High Resolution Imaging science Experiment) image to be processed back on Earth [11]. Image compression is one of the main methods that can reduce the problems based on removing redundant data. The image compression methods have two subcategories of lossless and lossy compression techniques [12, 13]. The differences between these two methods are that the image quality is high in lossless compression but compression ratio is less versus lossy compression [14, 15]. JPEG and JPEG2000 are two main methods which widely use in different fields [16-18]. The JPEG typically uses the discrete cosine transforms as main core in image compression. Also, The JPEG2000 is defined by the discrete wavelet transforms in order to reduce the volume of image space. In this work, I introduce a new image compression method based on gradient Haar wavelet [19, 20] which can be used in image compression standards for reducing the problems in human communications. The proposed method will showed the effect of sloping of scaling function of Haar wavelet on image compression. The statistical experimental results based on Mean Absolute Error (MAE), Peak Signal to Noise Ratio (PSNR), and Structural SIMilarity index (SSIM) show that the proposed method can improve the image compression standards.

This paper is organized as: an overview of related works discusses in section 2. I review the gradient Haar wavelet and then I introduce the preliminary image compression algorithm using gradient Haar wavelet transform in section 3. Then, I introduce a new image compression method for increasing of quality of image compression in section 4. Next, the statistical experimental results discuss in section 5, and finally in Section 6 provide a brief conclusion.

## II. OVERVIEW OF RELATED WORKS

In this section, the brief introductions are given to the JPEG and the JPEG2000 structures.

*A. JPEG*

Define The JPEG standard (Joint Photographic Experts Group that is one of the joint ISO/CCITT committees) is the first international image compression standard for still images (grayscale and color) [16]. The JPEG standard includes two basic compression methods (lossy and lossless methods) with various modes of operation that have many applications. The JPEG uses the discrete cosine transforms (DCT) for lossy image compression, although, there is a predictive method for lossless image compression (JPEG-LS). The JPEG processing includes block splitting, DCT, quantization, and entropy coding. Each block of image matrix is converted to a representation of the frequency domain based on the two dimensional DCT [21]. The method has high compression ratios and low image quality [22].

*B. JPEG2000*

The JPEG 2000 compression standard is the development of

The submitted date:.
Yaser Sadra is with the Department of computer sciences, Shandiz Institute of Higher Education, Mashhad, Iran (e-mail: y.sadra@shandiz.ac.ir).

a new standard for the compression of still images [23]. The JPEG 2000 presents both lossy and lossless image compressions. It proposes higher compression ratios for lossy image compression and also works on image tiles to save memory space. As above mentioned, the important part of the JPEG standard is the DCT, versus; the main part of the JPEG 2000 compression standard is the discrete wavelet transform (DWT) that is a method of the multi resolution processing [18, 24-26]. The DWT can obtain several approximations of a function at different levels of resolution which the wavelet function and the scaling function of wavelet are major functions in multi resolution processing [26, 27]. The DWT increases compression efficiency because of good capacity in coupling of the numbers and the ability to restore the numbers. Here, I use Haar wavelet transform as a the discrete wavelet transform in Jpeg2000. Nowadays, it is proposed as the substitution for the JPEG standard.

In simple terms, the JPEG 2000 processing first divide image into several non-overlapping tiles. Then, it accomplishes the DWT, quantization, and entropy coding on each tile that is similar to the JPEG [24]. Carefully on the DWT part, we know that the JPEG2000 algorithm uses the two dimensional DWT to decompose the original image to four sub images (LL, LH, HL and HH). The sub images are filtered with low pass (L) and high pass filters (H). The LL has been filtered using of low pass filters in horizontal and vertical directions that is the approximated matrix of original image. The HL has been filtered using of high pass filter in the vertical direction and low pass filter in horizontal direction that is a details matrix of original image. The LH has been filtered using of low pass filter in the vertical direction and high pass filter in horizontal direction which is a details matrix of original image. The HH has been filtered using of high pass filters in both horizontal and vertical directions which is a details matrix of original image [23, 24]. The LL image is quarter of the size of original image and can be decomposed to four new sub images.

### III. Gradient Haar Wavelet

In this section, I first review gradient Haar wavelet transform and then using them in a preliminary image compression algorithm.

*A. Gradient Haar wavelet transform*

The structure of Haar wavelet is simplest between wavelets. The property cause it use in the various sciences. The Haar wavelet was proposed in 1909 by Alfred Haar [28]. The gradient Haar wavelet (GHW) has been introduced in our pervious works [19, 20] which is a Haar wavelet with the sloping scaling function. The gradient Haar wavelet (GHW) can be more efficient than Haar wavelet due to sloping scaling function of it in signal processing. Because the main base of the signal processing of a wave based on wavelet is the scaling function. If the scaling function can better illustrate the wave in fewer repetitions, then, the signal processing of it has more information of the wave [19, 20]. To describe the gradient Haar wavelet (GHW) first considering scaling function of Haar wavelet that changed as a sloping step function

$$\phi(x) = \begin{cases} \gamma\left(x - \frac{1}{2}\right) + 1 & 0 \leq x < 1 \\ 0 & other\ wise \end{cases} \quad (1)$$

where $\gamma$ is line slop. Then, the Haar wavelet is the function:

$$\psi(x) = \begin{cases} (\frac{\gamma^2}{24} + \frac{\gamma}{4} + 1)(2\gamma x - \frac{\gamma}{2}+1) & 0 \leq x < \frac{1}{2} \\ -(\frac{\gamma^2}{24} - \frac{\gamma}{4} + 1)(2\gamma x - \frac{3\gamma}{2}+1) & \frac{1}{2} \leq x < 1 \\ 0 & otherwise \end{cases} \quad (2)$$

The other mathematical perspective, the scaling and wavelet functions of the GHW transform are as following

$$p_0 = \frac{\gamma^2}{24} - \frac{\gamma}{4} + 1,\ p_1 = \frac{\gamma^2}{24} + \frac{\gamma}{4} + 1 \quad (3)$$

$$\phi(x) = (\frac{\gamma^2}{24} - \frac{\gamma}{4} + 1)\phi(2x) + (\frac{\gamma^2}{24} + \frac{\gamma}{4} + 1)\phi(2x - 1)$$

$$\psi(x) = (\frac{\gamma^2}{24} + \frac{\gamma}{4} + 1)\phi(2x) - (\frac{\gamma^2}{24} - \frac{\gamma}{4} + 1)\phi(2x - 1)$$

where if the $\gamma$ is zero, the functions are the Haar wavelet functions. The GHW in two dimensions is obtained as similar to Haar wavelet. A $4 \times 4$ GHW transformation matrix is generated as follows

$$G_1 = \begin{vmatrix} \widetilde{p_0}(\gamma) & \widetilde{p_1}(\gamma) & 0 & 0 \\ 0 & 0 & \widetilde{p_0}(\gamma) & \widetilde{p_1}(\gamma) \\ \widetilde{p_1}(\gamma) & -\widetilde{p_0}(\gamma) & 0 & 0 \\ 0 & 0 & \widetilde{p_1}(\gamma) & -\widetilde{p_0}(\gamma) \end{vmatrix}\ and$$

$$G_2 = \begin{vmatrix} \widetilde{p_0}(\gamma) & \widetilde{p_1}(\gamma) & 0 & 0 \\ \widetilde{p_1}(\gamma) & -\widetilde{p_0}(\gamma) & 0 & 0 \\ 0 & 0 & 1 & 0 \\ 0 & 0 & 0 & 1 \end{vmatrix}$$

that $G_1$ and $G_2$ are first approximation matrix and Second approximation matrix, respectively. Then, we have a GHW transformation matrix with variable coefficients

$$G_{4 \times 4} = G_2 G_1 =$$
$$\begin{vmatrix} (\widetilde{p_0}(\gamma))^2 & \widetilde{p_0}(\gamma)\widetilde{p_1}(\gamma) & \widetilde{p_1}(\gamma)\widetilde{p_0}(\gamma) & (\widetilde{p_1}(\gamma))^2 \\ \widetilde{p_1}(\gamma)\widetilde{p_0}(\gamma) & (\widetilde{p_1}(\gamma))^2 & -(\widetilde{p_0}(\gamma))^2 & -\widetilde{p_0}(\gamma)\widetilde{p_1}(\gamma) \\ \widetilde{p_1}(\gamma) & -\widetilde{p_0}(\gamma) & 0 & 0 \\ 0 & 0 & \widetilde{p_1}(\gamma) & -\widetilde{p_0}(\gamma) \end{vmatrix} \quad (4)$$

where the $\widetilde{p_l}(\gamma) = \frac{p_l(\gamma)}{\sqrt{|p_0(\gamma)|^2 + |p_1(\gamma)|^2}}$ coefficients are scaling function coefficients. The GHW transformations could have many applications in applied sciences [19, 20].

*B. Preliminary image compression algorithm*

In this section, an image compression algorithm based on the GHW will be introduced. In the other word, I improve the DWT of JPEG2000 algorithm using GHW transform and to obtain the appropriate parameter ($\gamma$) for GHW transform. We can decompose the original image to four sub images (LL, LH, HL, and HH) that all of them are the homologous approximated-details matrix of original image. First, an original image is considered with $M_{m \times m}$ matrix. If G is be GHW transformation matrix, matrix E can be the first wavelet decomposition which contains the coefficients matrices LL, LH, HL, and HH of the original image. In order that all of coefficients matrices are be the approximated-details coefficients matrices, we should find the appropriate value of line slop '$\gamma$' of GHW. If coefficients matrices are defined as follows

$$\begin{bmatrix} LL \triangleq \begin{bmatrix} \widetilde{p_1}^2 x_{2i,2j} + \widetilde{p_1}\widetilde{p_0}(\alpha x_{2i,2j-1} + x_{2i-1,2j}) \\ +\widetilde{p_0}^2 x_{2i-1,2j-1} \end{bmatrix}_{1 \le i,j \le m} \\ LH \triangleq \begin{bmatrix} \widetilde{p_1}^2 x_{2i,2j-1} + \widetilde{p_1}\widetilde{p_0}(x_{2i-1,2j-1} - x_{2i,2j}) \\ -\widetilde{p_0}^2 x_{2i-1,2j} \end{bmatrix}_{1 \le i,j \le m} \\ HL \triangleq \begin{bmatrix} \widetilde{p_1}^2 x_{2i-1,2j} + \widetilde{p_1}\widetilde{p_0}(x_{2i-1,2j-1} - x_{2i,2j}) \\ -\widetilde{p_0}^2 x_{2i,2j-1} \end{bmatrix}_{1 \le i,j \le m} \\ HH \triangleq \begin{bmatrix} \widetilde{p_1}^2 x_{2i-1,2j-1} - \widetilde{p_1}\widetilde{p_0}(x_{2i,2j-1} + x_{2i-1,2j}) \\ +\alpha\widetilde{p_0}^2 x_{2i,2j} \end{bmatrix}_{1 \le i,j \le m} \end{bmatrix} \quad (5)$$

where $\alpha$ is a constant value. The total coefficients of matrix E is be as following

$$S = \sum_{i,j=1}^{m} x_{i,j} \quad (6)$$

In result, total coefficients of matrices of sub images as following

$$\begin{bmatrix} S_{LL} = \widetilde{p_1}^2 \sum_{i,j=1}^{m} x_{2i,2j} + \\ \widetilde{p_1}\widetilde{p_0} \sum_{i,j=1}^{m}(\alpha x_{2i,2j-1} + x_{2i-1,2j}) + \widetilde{p_0}^2 \sum_{i,j=1}^{m} x_{2i-1,2j-1} \\ S_{LH} = \widetilde{p_1}^2 \sum_{i,j=1}^{m} x_{2i,2j-1} + \\ \widetilde{p_1}\widetilde{p_0} \sum_{i,j=1}^{m}(x_{2i-1,2j-1} - x_{2i,2j}) - \widetilde{p_0}^2 \sum_{i,j=1}^{m} x_{2i-1,2j} \\ S_{HL} = \widetilde{p_1}^2 \sum_{i,j=1}^{m} x_{2i-1,2j} + \\ \widetilde{p_1}\widetilde{p_0} \sum_{i,j=1}^{m}(x_{2i-1,2j-1} - x_{2i,2j}) - \widetilde{p_0}^2 \sum_{i,j=1}^{m} x_{2i,2j-1} \\ S_{HH} = \widetilde{p_1}^2 \sum_{i,j=1}^{m} x_{2i-1,2j-1} - \\ \widetilde{p_1}\widetilde{p_0} \sum_{i,j=1}^{m}(x_{2i,2j-1} + x_{2i-1,2j}) + \alpha\widetilde{p_0}^2 \sum_{i,j=1}^{m} x_{2i,2j} \end{bmatrix} \quad (7)$$

In the normal mode, total coefficients of matrices of sub images (LL, LH, HL, and HH) are not equal. For example, if we consider the difference between two total coefficients of matrices of sub image matrices LH and HL which is equal to a fixed value $k$, i.e.

$$S_{LH} = S_{HL} + k \quad (8)$$

Then,

$$\widetilde{p_1}^2 C + \widetilde{p_1}\widetilde{p_0} B - \widetilde{p_0}^2 A = \widetilde{p_1}^2 A + \widetilde{p_1}\widetilde{p_0} B - \widetilde{p_0}^2 C + k \quad (9)$$

where A, B, C and D are as following

$$A = \sum_{i,j=1}^{m} x_{2i-1,2j}, \quad B = \sum_{i,j=1}^{m}(x_{2i-1,2j-1} - x_{2i,2j}),$$
$$C = \sum_{i,j=1}^{m} x_{2i,2j-1}.$$

The Eq. 9 can be summarized as follows

$$\widetilde{p_1}^2 + \widetilde{p_0}^2 = \frac{k}{C-A}. \quad (10)$$

By substituting Eq. 3 in the Eq. 10, we have

$$\gamma = \sqrt{6 \frac{\sqrt{(C-A)(33(C-A)+8k)} - 7(C-A)}{C-A}}$$

Since we want the sub images are be same then the value of k should be equal to zero (k=0), in result,

$$\gamma = \sqrt{6\sqrt{33} - 42} = 2.7445626465380280\, i$$

The values of $\widetilde{p_0}$ and $\widetilde{p_1}$ using of the complex value of $\gamma$ are as following

$$\begin{cases} \widetilde{p_0} = \frac{p_0}{\sqrt{|p_0|^2 + |p_1|^2}} = 0.5(1 - i) \\ \widetilde{p_1} = \frac{p_1}{\sqrt{|p_0|^2 + |p_1|^2}} = 0.5(1 + i) \end{cases} \quad (11)$$

This property of GHW which the values of $\widetilde{p_0}$ and $\widetilde{p_1}$ can have a wide range of numbers depending on the application, is the superiority of it is relative to Haar wavelet. Therefore, we can carry out image compression using the values of $\widetilde{p_0}$ and $\widetilde{p_1}$ and based on GHW transformation, i.e.

$$E = G^T M G. \quad (12)$$

To consider $\alpha = -1$, elements of the matrix E are as follows

$$\begin{bmatrix} LL \triangleq \left[ \frac{x_{2i-1,2j} - x_{2i,2j-1}}{2} - \frac{x_{2i-1,2j-1} - x_{2i,2j}}{2} i \right]_{1 \le i,j \le m} \\ LH \triangleq \left[ \frac{x_{2i-1,2j-1} - x_{2i,2j}}{2} + \frac{x_{2i-1,2j} + x_{2i,2j-1}}{2} i \right]_{1 \le i,j \le m} \\ HL \triangleq \left[ \frac{x_{2i-1,2j-1} - x_{2i,2j}}{2} + \frac{x_{2i-1,2j} + x_{2i,2j-1}}{2} i \right]_{1 \le i,j \le m} \\ HH \triangleq \left[ -\frac{x_{2i-1,2j} + x_{2i,2j-1}}{2} + \frac{x_{2i-1,2j-1} + x_{2i,2j}}{2} i \right]_{1 \le i,j \le m} \end{bmatrix} \quad (13)$$

where the coefficients matrix E are complex values. In result, we can obtain modulus of the coefficients matrix E as follows

$$F_{i,j} = |E_{i,j}| \quad (14)$$

that | | is symbol of absolute value. Hence, we have four identical sub images that all of them are the approximated-details coefficients matrices of the original image (see fig 1).

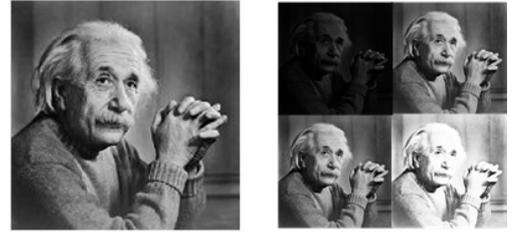

Fig.1. (a) The original image ''Einstein''. (b) The modulus of the coefficients matrices of 1-level wavelet decomposition of the image using the gradient Haar wavelet.

Each sub images can be used for the 2nd level of the wavelet decomposition. Also, they equally inherit the features of the original image.

IV. THE PROPOSED METHOD

In this section, I propose a new image compression method based on preliminary image compression algorithm that in addition to above advantages, it has much greater flexibility in data saving. For simplicity in the description of the method, I explain it for a gray image that can easily be generalized to color images. If we look more closely GHW transform, first it divides each matrix of image into blocks of $2 \times 2$ and then performs computations on the blocks. Here, we can use the property to develop preliminary image compression algorithm to a new image compression method.

## A. Compression process

For simplicity in the description of the algorithm, first, a original image is considered with $M_{m \times m}$ matrix that m is even. The algorithm can be used to $M_{m \times n}$ matrices that m and n are even and also $m \neq n$. Second, matrix of averages of each block is computed from Eq. 13 as following

$$S \triangleq [\|\,|Re(HH)| + |Im(HH)|\,\|] = \left[\left\|\frac{x_{2i-1,2j-1}+x_{2i,2j}+x_{2i-1,2j}+x_{2i,2j-1}}{4}\right\|\right]_{1 \leq i,j \leq m} \quad (15)$$

that $\|\ \|$ is a round bracket. Then, matrices of differences between elements of the primary diagonal ($N_1$) diagonal and secondary diagonal ($N_2$) and also absolute values $N_1$ and $N_2$ are computed based on Eq. 13 as follows

$$\begin{cases} N_1 \triangleq [-2\,Im(LL)]_{1 \leq i,j \leq m} \\ \quad [x_{2i-1,2j-1} - x_{2i,2j}]_{1 \leq i,j \leq m} \\ N_2 \triangleq [2\,Re(LL)]_{1 \leq i,j \leq m} \\ \quad [x_{2i-1,2j} - x_{2i,2j-1}]_{1 \leq i,j \leq m} \\ N_3 \triangleq [|N_1| - |N_2|]_{1 \leq i,j \leq m}. \end{cases} \quad (16)$$

To use matrices of differences, we can compute matrix of maximum distances between elements of diagonals as following

$$A \triangleq \left[\begin{cases} |N_1| + |N_2| + |N_3|, & (N_1 \geq N_2 \ \& \ N_1 \geq 0) \\ & \oplus (N_1 < N_2 \ \& \ N_2 \geq 0) \\ -(|N_1| + |N_2| + |N_3|), & (N_1 \geq N_2 \ \& \ N_1 < 0) \\ & \oplus (N_1 < N_2 \ \& \ N_2 < 0) \end{cases}\right]_{1 \leq i,j \leq m} \quad (17)$$

that their values can be positive or negative. In result, we can compute coefficients matrices LH and HL as following

$$\overline{LH} \triangleq \begin{bmatrix} [\lceil|S + \lambda \times A|\rceil], & A \geq 0 \\ [\lfloor|S + \lambda \times A|\rfloor], & A < 0 \end{bmatrix}_{1 \leq i,j \leq m} \quad (18)$$

$$\overline{HL} \triangleq \begin{bmatrix} [\lfloor|S - \lambda \times A|\rfloor], & A \geq 0 \\ [\lceil|S - \lambda \times A|\rceil], & A < 0 \end{bmatrix}_{1 \leq i,j \leq m} \quad (19)$$

where $\lambda$ is a balancing coefficient that can be equal to $\frac{1}{8}$ and also $\lceil\ \rceil$ $and$ $\lfloor\ \rfloor$ are ceiling brackets and floor brackets, respectively. With this method, the LH and the HL can also save position information of elements of blocks for the diagonal that has maximum difference between elements (major diagonal). If the $[\overline{lh}_{i,j}]$ is greater than or equal to the $[\overline{hl}_{i,j}]$, it means that the up element of the major diagonal has larger number. And if the $[\overline{lh}_{i,j}]$ is less than the $[\overline{hl}_{i,j}]$, it means that the down element of the major diagonal has larger number. Consequently, the coefficients matrices LL and HH can be zero

$$\begin{cases} LL \triangleq [0]_{1 \leq i,j \leq m} \\ HH \triangleq [0]_{1 \leq i,j \leq m} \end{cases}. \quad (20)$$

In order to increase position information of elements of blocks, we use two masks that their sizes are the same the LH and the HL and are defined as follows

$$B \triangleq \left[\begin{cases} even, & N_1 \geq N_2 \ \& \ N_2 \geq 0 \\ odd, & N_1 \geq N_2 \ \& \ N_2 < 0 \\ even, & N_1 < N_2 \ \& \ N_1 \geq 0 \\ odd, & N_1 < N_2 \ \& \ N_1 < 0 \end{cases}\right]_{1 \leq i,j \leq m} \quad (21)$$

that the mask B saves position information of the diagonal that has minimum difference between elements (minor diagonal). If the $[b_{i,j}]$ is even, it means that the up element of the minor diagonal has larger number. And if the $[b_{i,j}]$ is odd, it means that the down element of the minor diagonal has larger number. Here, to note that major and minor diagonals are not mean primary and secondary diagonals.

$$C \triangleq \left[\begin{cases} even, & (x_{2i-1,2j-1} + x_{2i,2j-1}) \geq (x_{2i,2j} + x_{2i-1,2j}) \\ odd, & (x_{2i-1,2j-1} + x_{2i,2j-1}) < (x_{2i,2j} + x_{2i-1,2j}) \end{cases}\right]_{1 \leq i,j \leq m} \quad (22)$$

that the mask C saves position information of columns. If the $[c_{i,j}]$ is even, it means that the left column of the block is greater than or equal to the right column. And if the $[c_{i,j}]$ is odd, it means that the left column of the block is less than the right column.

Finally, we filter mask B on the HL and mask C on the LH. In result, we have two details-approximation coefficients matrices that in total are half of size of the matrix of the original image.

$$E \triangleq \begin{bmatrix} 0 & [\![HL]\!]_B \\ [\![LH]\!]_C & 0 \end{bmatrix} \quad (23)$$

that $[\![\ ]\!]$ is symbol of masking. The proposed method has the ability to be combined with other compression methods. An example of a 4 × 4 matrix is shown in follows

$$M_{4 \times 4} = \begin{bmatrix} 61 & 69 & 79 & 67 \\ 59 & 67 & 81 & 72 \\ 54 & 60 & 74 & 60 \\ 55 & 63 & 61 & 34 \end{bmatrix}.$$

The matrix of averages of each block is as following

$$S_{2 \times 2} = \begin{bmatrix} 64 & 75 \\ 58 & 57 \end{bmatrix}.$$

And also, matrices of differences between diagonal elements are as following

$$N_{1_{2 \times 2}} = \begin{bmatrix} -6 & 7 \\ -9 & 40 \end{bmatrix}, N_{2_{2 \times 2}} = \begin{bmatrix} 10 & -14 \\ 5 & -1 \end{bmatrix},$$

$$N_{3_{2 \times 2}} = \begin{bmatrix} -4 & -7 \\ 4 & 39 \end{bmatrix}.$$

Therefore, the matrix of maximum distances between diagonal elements is as following

$$A_{2 \times 2} = \begin{bmatrix} 20 & -28 \\ -18 & 80 \end{bmatrix}.$$

In result, coefficients matrices LH and HL with the balancing coefficient $\lambda = \frac{1}{8}$ are as following

$$\overline{LH}_{2 \times 2} = \begin{bmatrix} 67 & 71 \\ 55 & 67 \end{bmatrix}, \overline{HL}_{2 \times 2} = \begin{bmatrix} 61 & 79 \\ 61 & 47 \end{bmatrix}.$$

The mask B and the mask C that save position information of elements of blocks can be obtained as follows

$$B_{2 \times 2} = \begin{bmatrix} 1 & 0 \\ 0 & 1 \end{bmatrix}, C_{2 \times 2} = \begin{bmatrix} 1 & 0 \\ 1 & 0 \end{bmatrix}$$

where numbers zero and one are symbol of even and odd, respectively. Finally, to filter mask B on the HL and mask C

on the LH, we have compressed matrix E as following

$$E_{4\times 4} = \begin{bmatrix} 0 & 0 & 61 & 78 \\ 0 & 0 & 60 & 47 \\ 67 & 70 & 0 & 0 \\ 55 & 66 & 0 & 0 \end{bmatrix}.$$

By comparing the matrix E and M, we see that the volume of Information of matrix E is half the matrix M. The block diagram of the proposed compressed method is displayed in fig 2.

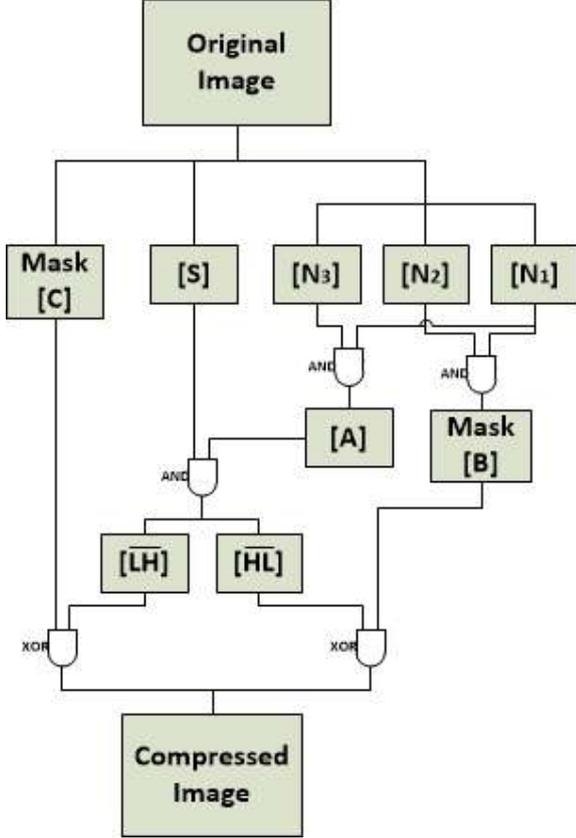

Fig. 2. The block diagram of the proposed compressed method.

### B. Decompression process

For The decompression process can be almost the same as the compression process but with reverse steps. For more explanation, we describe it in following. First, a compressed image (that was compressed by proposed method) is considered with matrix $E_{m\times m}$ that m is even. Then, we can define matrix $Á$ for each block from $[\![\overline{LH}]\!]_C$ and $[\![\overline{HL}]\!]_B$.

$$Á \triangleq \begin{bmatrix} \acute{x}_{2i-1,2j-1} & \acute{x}_{2i-1,2j} \\ \acute{x}_{2i,2j-1} & \acute{x}_{2i,2j} \end{bmatrix}_{1\le i,j\le m} \quad (24)$$

and the altered state of matrix $Á$ can be as follows

$$Á^T \triangleq \begin{bmatrix} \acute{x}_{2i-1,2j} & \acute{x}_{2i-1,2j-1} \\ \acute{x}_{2i,2j} & \acute{x}_{2i,2j-1} \end{bmatrix}_{1\le i,j\le m} \quad (25)$$

The elements of the matrix $Á$ are obtained from comparing the $\overline{LH}$ to the $\overline{HL}$ as follows

$$\acute{x}_{2i-1,2j-1} \triangleq$$

$$\left[ \begin{cases} \frac{\overline{lh}_{i,j}+\overline{hl}_{i,j}}{2} + \mu \times |\overline{lh}_{i,j}-\overline{hl}_{i,j}|, & \overline{lh}_{i,j} \ge \overline{hl}_{i,j} \\ \frac{\overline{lh}_{i,j}+\overline{hl}_{i,j}}{2} - \mu \times |\overline{lh}_{i,j}-\overline{hl}_{i,j}|, & \overline{lh}_{i,j} < \overline{hl}_{i,j} \end{cases} \right]_{1\le i,j\le m} \quad (26)$$

$$\acute{x}_{2i-1,2j} \triangleq$$

$$\left[ \begin{cases} \overline{lh}_{i,j}, & (\overline{hl}_{i,j} \ge \overline{lh}_{i,j} \ \& \ \overline{hl}_{i,j} \text{ be odd}) \\ & \oplus (\overline{hl}_{i,j} < \overline{lh}_{i,j} \ \& \ \overline{hl}_{i,j} \text{ be even}) \\ \overline{hl}_{i,j}, & (\overline{hl}_{i,j} < \overline{lh}_{i,j} \ \& \ \overline{hl}_{i,j} \text{ be odd}) \\ & \oplus (\overline{hl}_{i,j} \ge \overline{lh}_{i,j} \ \& \ \overline{hl}_{i,j} \text{ be even}) \end{cases} \right]_{1\le i,j\le m} \quad (27)$$

$$\acute{x}_{2i,2j-1} \triangleq$$

$$\left[ \begin{cases} \overline{hl}_{i,j}, & (\overline{hl}_{i,j} \ge \overline{lh}_{i,j} \ \& \ \overline{hl}_{i,j} \text{ be odd}) \\ & \oplus (\overline{hl}_{i,j} < \overline{lh}_{i,j} \ \& \ \overline{hl}_{i,j} \text{ be even}) \\ \overline{lh}_{i,j}, & (\overline{hl}_{i,j} < \overline{lh}_{i,j} \ \& \ \overline{hl}_{i,j} \text{ be odd}) \\ & \oplus (\overline{hl}_{i,j} \ge \overline{lh}_{i,j} \ \& \ \overline{hl}_{i,j} \text{ be even}) \end{cases} \right]_{1\le i,j\le m} \quad (28)$$

and also

$$\acute{x}_{2i,2j} \triangleq$$

$$\left[ \begin{cases} \frac{\overline{lh}_{i,j}+\overline{hl}_{i,j}}{2} - \mu \times |\overline{lh}_{i,j}-\overline{hl}_{i,j}|, & \overline{lh}_{i,j} \ge \overline{hl}_{i,j} \\ \frac{\overline{lh}_{i,j}+\overline{hl}_{i,j}}{2} + \mu \times |\overline{lh}_{i,j}-\overline{hl}_{i,j}|, & \overline{lh}_{i,j} < \overline{hl}_{i,j} \end{cases} \right]_{1\le i,j\le m} \quad (29)$$

where $\mu$ is a recursive balancing coefficient that can be equal to 0.97. If we can define matrices $\acute{N}_1$ and $\acute{N}_2$ as following

$$\begin{cases} \acute{N}_1 \triangleq [\acute{x}_{2i-1,2j-1} + \acute{x}_{2i,2j-1}]_{1\le i,j\le m} \\ \acute{N}_2 \triangleq [\acute{x}_{2i,2j} + \acute{x}_{2i-1,2j}]_{1\le i,j\le m} \end{cases} \quad (30)$$

the matrix $\acute{B}$ is obtained from the $\acute{N}_1$, the $\acute{N}_2$ and the $\overline{LH}$ as follows

$$\acute{B} \triangleq$$

$$\left[ \begin{cases} Á, & \left(\acute{n}_{1_{i,j}} < \acute{n}_{2_{i,j}} \ \& \ lh_{i,j} \text{ be odd}\right) \\ & \oplus \left(\acute{n}_{1_{i,j}} \ge \acute{n}_{2_{i,j}} \ \& \ lh_{i,j} \text{ be even}\right) \\ Á^T, & \left(\acute{n}_{1_{i,j}} \ge \acute{n}_{2_{i,j}} \ \& \ lh_{i,j} \text{ be odd}\right) \\ & \oplus \left(\acute{n}_{1_{i,j}} < \acute{n}_{2_{i,j}} \ \& \ lh_{i,j} \text{ be even}\right) \end{cases} \right]_{1\le i,j\le m} \quad (31)$$

In result, the decompressed matrix $\acute{M}$ is obtained as follows

$$\acute{M} \triangleq [\|\acute{B}\|]_{1\le i,j\le m} \quad (32)$$

that $\| \ \|$ is a round bracket. The above example can be used to decompressed process as follows

$$E_{4\times 4} = \begin{bmatrix} 0 & 0 & 61 & 78 \\ 0 & 0 & 60 & 47 \\ 67 & 70 & 0 & 0 \\ 55 & 66 & 0 & 0 \end{bmatrix}.$$

The matrix $Á$ and the altered state of it $Á^T$ are obtained as follows

$$Á_{4\times 4} = \begin{bmatrix} 70 & 61 & 66 & 78 \\ 67 & 58 & 70 & 82 \\ 52.5 & 60 & 75.5 & 47 \\ 55 & 62.5 & 66 & 37.5 \end{bmatrix},$$

$$\acute{A}^T{}_{4\times 4} = \begin{bmatrix} 61 & 70 & 78 & 66 \\ 58 & 67 & 82 & 70 \\ 60 & 52.5 & 47 & 75.5 \\ 62.5 & 55 & 37.5 & 66 \end{bmatrix}.$$

where the recursive balancing coefficient is $\mu = 0.97$. In result, the matrix $\acute{B}$ is obtained as follows

$$\acute{B}_{4\times 4} = \begin{bmatrix} 61 & 70 & 78 & 66 \\ 58 & 67 & 82 & 70 \\ 52.5 & 60 & 75.5 & 47 \\ 55 & 62.5 & 66 & 37.5 \end{bmatrix}.$$

Finally, the decompressed matrix $\acute{M}$ is obtained as follows

$$\acute{M}_{4\times 4} = \begin{bmatrix} 61 & 70 & 78 & 66 \\ 58 & 67 & 82 & 70 \\ 53 & 60 & 76 & 47 \\ 55 & 63 & 66 & 38 \end{bmatrix}.$$

A mean absolute error (MAE) between $M_{4\times 4}$ and $\acute{M}_{4\times 4}$ is equal to 2. Whatever the matrix size is larger, the MAE is closely to zero.

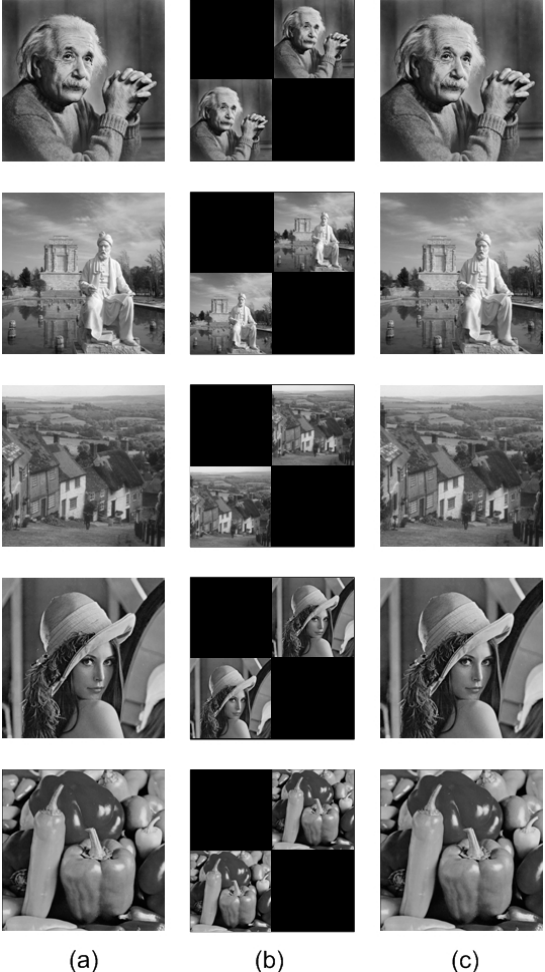

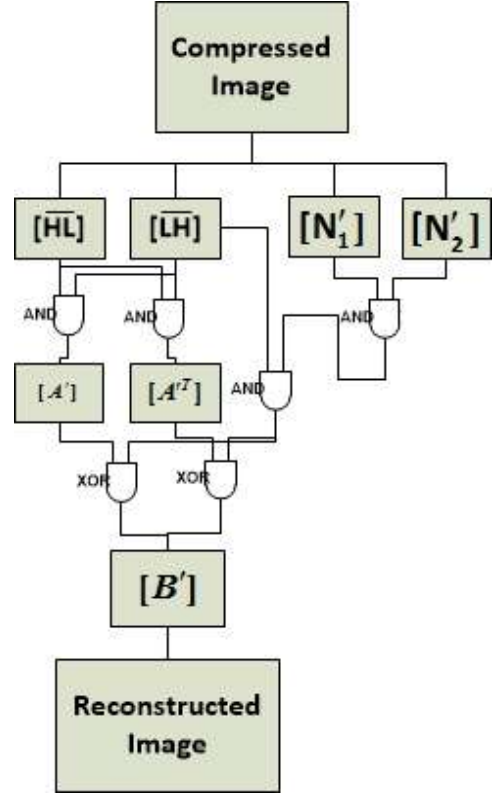

Fig. 4. The block diagram of the proposed decompressed method.

The proposed method can has more iteration, that is, we can several times iterate the algorithm on sub images. Also, the algorithm can be combined with other coding and algorithms to increase the compression rate.

## V. IMAGE QUALITY MEASURE AND STATISTICAL ANALYSIS

The proposed method is simulated on a moderate system (Intel i3 processor with 4 GB RAM) based on MATLAB[TM] version 7.10 and is tested on standard 8-bit/pixel grayscale bitmap images with size of $512 \times 512$ pixels which have been updated to Google Drive:

(https://drive.google.com/drive/folders/0B7j_o4refVy3RG5lZ25veDFmMms). There are many parametric measures to examine the quality of an image such as Compression Ratio (CR), Mean Absolute Error (MAE), Peak Signal to Noise Ratio (PSNR), and Structural SIMilarity index (SSIM). The CR is the ratio of uncompressed size $|M|$ of the original image to compressed size $|E|$ of the original image, i.e. $CR = \frac{|M|}{|E|}$. The MAE is defined as follows

$$MAE = \frac{1}{m \times n} \sum_{i=1}^{m} \sum_{j=1}^{n} |M(i,j) - \acute{M}(i,j)| \qquad (33)$$

Fig.3. (a) The original images ''Einstein, Ferdowsi, Goldhill, Lena, Peppers''. (b) The details-approximation coefficients matrices of 1-level decomposition of the images using the proposed compressed method. (c) The reconstructed images using the proposed decompressed method.

The fig 3 illustrates the steps of proposed method on original images. The block diagram of the proposed decompressed method is displayed in fig 4.

that the $M(i,j)$ and $\acute{M}(i,j)$ are the pixels of the original image and the reconstructed image, respectively. If the $M(i,j)$ and $\acute{M}(i,j)$ are very close to each other, the MAE can be closely to zero. The PSNR is the ratio between the maximum possible power of a signal and the point wise difference between the original image and the reconstructed image which is represented as following equation

$$PSNR = 10 \log_{10} \frac{255^2}{MSE} \quad (34)$$

that Mean Square Error (MSE) is as

$$MSE = \frac{1}{m \times n} \sum_{i=1}^{m} \sum_{j=1}^{n} \left(M(i,j) - \acute{M}(i,j)\right)^2 \quad (35)$$

The range of PSNR is between 0 and 100 dB. Good reconstructed images usually have PSNR values more than 30 dB. The SSIM index determines the similarity between original and reconstructed images [29]. Also, the SSIM index can examine the image resolution using a plain reference image. The SSIM index is defined as follows [30]

$$SSIM(m, \acute{m}) = \frac{(2\mu_m \mu_{\acute{m}} + C_1)(2\sigma_{m\acute{m}} + C_2)}{(\mu_m^2 + \mu_{\acute{m}}^2 + C_1)(\sigma_m^2 + \sigma_{\acute{m}}^2 + C_2)} \quad (36)$$

where $\mu_m, \mu_{\acute{m}}, \sigma_m^2, \sigma_{\acute{m}}^2$ and $\sigma_{m\acute{m}}$ are averages, variances and covariance of $m$ and $\acute{m}$, respectively. And also, $\{C_i\}_{i=1,2}$ are two variables to stabilize the division with weak denominator which are obtained as $C_i = (k_i L)^2$ with $L = 255, k_1 = 0.01$ and $k_2 = 0.03$ [29].

Here, in order to show the abilities of proposed method, we use proposed method without quantization and entropy coding. Therefore, it has been compared to the original lossy modes of JPEG and JPEG2000 (based on Haar wavelet) methods.

TABLE I
MAE RESULTS OF RECONSTRUCTED IMAGES FOR CR=2

| Image | JPEG | JPEG2000 | Proposed method |
|---|---|---|---|
| Einstein | 2.3618 | 2.6810 | 1.2185 |
| Ferdowsi | 2.4801 | 2.3107 | 0.9816 |
| Goldhill | 2.7583 | 2.4829 | 1.0873 |
| Lena | 2.1679 | 2.0683 | 0.8206 |
| Peppers | 2.3180 | 2.1685 | 1.0951 |

TABLE II
MAE RESULTS OF RECONSTRUCTED IMAGES FOR CR=4

| Image | JPEG | JPEG2000 | Proposed method |
|---|---|---|---|
| Einstein | 5.2584 | 5.3143 | 2.5615 |
| Ferdowsi | 4.8146 | 4.6285 | 2.0012 |
| Goldhill | 5.3483 | 5.0963 | 2.2361 |
| Lena | 4.3791 | 4.1465 | 1.7983 |
| Peppers | 4.4567 | 4.2624 | 2.2328 |

TABLE III
MAE RESULTS OF RECONSTRUCTED IMAGES FOR CR=8

| Image | JPEG | JPEG2000 | Proposed method |
|---|---|---|---|
| Einstein | 9.7849 | 9.8457 | 4.8816 |
| Ferdowsi | 9.3103 | 9.0684 | 4.0864 |
| Goldhill | 9.9576 | 9.5981 | 4.3943 |
| Lena | 8.4180 | 8.1763 | 3.6402 |
| Peppers | 8.9962 | 8.3475 | 4.3726 |

As we know, MAE is a scale-dependent accuracy measure which shows the accuracy of restored values [31]. Tables 1, 2, and 3 are presenting the quantitative comparison between JPEG, JPEG2000 (based on 1-level, 2-level, and 3-level decompositions) and proposed method (based on 1-level, 2-level, and 3-level decompositions) using MAE for Compression Ratios of 2, 4, and 8. Results show in compression ratios, the MAE values of proposed method are low than the half of JPEG and JPEG2000 methods.

TABLE IV
PSNR RESULTS OF RECONSTRUCTED IMAGES FOR CR=2

| Image | JPEG | JPEG2000 | Proposed method |
|---|---|---|---|
| Einstein | 28.3691 | 34.7503 | 41.5819 |
| Ferdowsi | 29.3857 | 35.1387 | 43.1670 |
| Goldhill | 31.3890 | 36.2751 | 44.1538 |
| Lena | 31.7902 | 36.4950 | 44.8604 |
| Peppers | 31.5089 | 35.8265 | 43.5937 |

TABLE V
PSNR RESULTS OF RECONSTRUCTED IMAGES FOR CR=4

| Image | JPEG | JPEG2000 | Proposed method |
|---|---|---|---|
| Einstein | 23.1684 | 28.4831 | 35.1356 |
| Ferdowsi | 24.0059 | 29.0500 | 37.2223 |
| Goldhill | 26.2347 | 30.3772 | 38.0521 |
| Lena | 26.5468 | 30.5374 | 38.6114 |
| Peppers | 26.4261 | 29.9899 | 37.4348 |

TABLE VI
PSNR RESULTS OF RECONSTRUCTED IMAGES FOR CR=8

| Image | JPEG | JPEG2000 | Proposed method |
|---|---|---|---|
| Einstein | 19.2350 | 28.4831 | 30.1286 |
| Ferdowsi | 20.3191 | 29.0500 | 32.4086 |
| Goldhill | 22.0019 | 30.3772 | 33.2189 |
| Lena | 22.4831 | 30.5374 | 33.5280 |
| Peppers | 22.3890 | 29.9899 | 32.5178 |

The PSNR is an approximation to human perception of reconstruction quality [32]. In Tables 4, 5, and 6 we see that the PSNR values show the performance of the proposed method transcends JPEG and JPEG2000 methods.

TABLE VII
SSIM RESULTS OF RECONSTRUCTED IMAGES FOR CR=2

| Image | JPEG | JPEG2000 | Proposed method |
|---|---|---|---|
| Einstein | 0.9553 | 0.9768 | 0.9940 |
| Ferdowsi | 0.9637 | 0.9829 | 0.9961 |
| Goldhill | 0.9675 | 0.9843 | 0.9968 |
| Lena | 0.9701 | 0.9899 | 0.9989 |
| Peppers | 0.9762 | 0.9928 | 0.9997 |

TABLE VIII
SSIM RESULTS OF RECONSTRUCTED IMAGES FOR CR=4

| Image | JPEG | JPEG2000 | Proposed method |
|---|---|---|---|
| Einstein | 0.9468 | 0.9674 | 0.9833 |
| Ferdowsi | 0.9542 | 0.9714 | 0.9857 |
| Goldhill | 0.9567 | 0.9736 | 0.9860 |
| Lena | 0.9608 | 0.9792 | 0.9881 |
| Peppers | 0.9673 | 0.9812 | 0.9894 |

TABLE IX
SSIM RESULTS OF RECONSTRUCTED IMAGES FOR CR=8

| Image | JPEG | JPEG2000 | Proposed method |
|---|---|---|---|
| Einstein | 0.9359 | 0.9561 | 0.9728 |

| Image | | | |
|---|---|---|---|
| Ferdowsi | 0.9430 | 0.9601 | 0.9749 |
| Goldhill | 0.9458 | 0.9620 | 0.9751 |
| Lena | 0.9502 | 0.9689 | 0.9775 |
| Peppers | 0.9569 | 0.9719 | 0.9786 |

The SSIM values of proposed method are compared with the SSIM values of JPEG and JPEG2000 methods in tables 7, 8, and 9. The results show that the proposed method improves the image quality of reconstructed images compared with the other two methods.

TABLE X
RUNNING TIME RESULTS OF RECONSTRUCTED IMAGES FOR CR=4

| Image | JPEG | JPEG2000 | Proposed method |
|---|---|---|---|
| Einstein | 1.28 | 0.76 | 0.68 |
| Ferdowsi | 1.32 | 0.71 | 0.61 |
| Goldhill | 1.37 | 0.74 | 0.65 |
| Lena | 1.15 | 0.66 | 0.59 |
| Peppers | 1.27 | 0.69 | 0.63 |

Table X presents the average run time for JPEG, JPEG2000 (based on 2-level decomposition) and proposed method (based on 2-level decomposition) in CR=4. Compared with JPEG, JPEG2000 implemented with MATLAB, the proposed method needs about 0.63s to process one image on average, which is faster than JPEG2000 (0.71s) and JPEG (1.27s).

In general, the statistical results show that the proposed method is a lossy compression method with high performance. Also, the results are shown that the proposed method can raise quality of image compression techniques if image compression standards are based on it.

## VI. CONCLUSION

In this paper, I have proposed a new image compression method based on gradient Haar wavelet. The proposed method was using gradient Haar wavelet transform in order to increase image quality of reconstructed image and decrease size of compressed image, simultaneously. Also, the results of the proposed method were compared with the JPEG and JPEG2000 (based on Haar wavelet) methods. The statistical results showed the advantage of the proposed method is higher than the other same level methods. Therefore, I propose the proposed method is used in image compression standards.

REFERENCES

1. M. N. Do, Q. H. Nguyen, H. T. Nguyen, D. Kubacki and S. J. Patel, "Immersive Visual Communication," in *IEEE Signal Processing Magazine*, vol. 28, no. 1, pp. 58-66, Jan. 2011.
2. Fuzheng Yang, Shuai Wan, "Bitstream-based quality assessment for networked video: a review", *Communications Magazine IEEE*, vol. 50, 2012, ISSN 0163-6804.
3. A. Borji and L. Itti, "State-of-the-Art in Visual Attention Modeling," in *IEEE Transactions on Pattern Analysis and Machine Intelligence*, vol. 35, no. 1, pp. 185-207, Jan. 2013.
4. K. Ma, H. Yeganeh, K. Zeng and Z. Wang, "High Dynamic Range Image Compression by Optimizing Tone Mapped Image Quality Index," in *IEEE Transactions on Image Processing*, vol. 24, no. 10, pp. 3086-3097, Oct. 2015.
5. A. Taneja, L. Ballan and M. Pollefeys, "Geometric Change Detection in Urban Environments Using Images," in *IEEE Transactions on Pattern Analysis and Machine Intelligence*, vol. 37, no. 11, pp. 2193-2206, Nov. 1 2015.
6. H. Nejati, V. Pomponiu, T. T. Do, Y. Zhou, S. Iravani and N. M. Cheung, "Smartphone and Mobile Image Processing for Assisted Living: Health-monitoring apps powered by advanced mobile imaging algorithms," in *IEEE Signal Processing Magazine*, vol. 33, no. 4, pp. 30-48, July 2016.
7. R. Wang *et al.*, "The MPEG Internet Video-Coding Standard [Standards in a Nutshell]," in *IEEE Signal Processing Magazine*, vol. 33, no. 5, pp. 164-172, Sept. 2016.
8. J. Portilla, V. Strela, M. J. Wainwright and E. P. Simoncelli, "Image denoising using scale mixtures of Gaussians in the wavelet domain," in *IEEE Transactions on Image Processing*, vol. 12, no. 11, pp. 1338-1351, Nov. 2003.
9. Y. Kwon, K. I. Kim, J. Tompkin, J. H. Kim and C. Theobalt, "Efficient Learning of Image Super-Resolution and Compression Artifact Removal with Semi-Local Gaussian Processes," in *IEEE Transactions on Pattern Analysis and Machine Intelligence*, vol. 37, no. 9, pp. 1792-1805, Sept. 1 2015.
10. B. E. Usevitch, "A tutorial on modern lossy wavelet image compression: foundations of JPEG 2000," in *IEEE Signal Processing Magazine*, vol. 18, no. 5, pp. 22-35, Sep 2001.
11. J. Yuen, L. Deutsch and S. Townes, "Deep Space Communications," in JPL Publication 400-1385. Jet Propulsion Laboratory, California Institute of Technology (Pasadena, CA), 2017, (https://scienceandtechnology.jpl.nasa.gov/research/research-topics-list/communications-computing-software/deep-space-communications).
12. A. M. Rufai, Gh. Anbarjafari, H. Demirel, "Lossy image compression using singular value decomposition and wavelet dif ference re duction, " in Digital Signal Processing, vol. 24, pp. 117-123, Sep 2103.
13. N. Goel and S. Gabarda, "Lossy and lossless image compression using Legendre polynomials," *2016 Conference on Advances in Signal Processing (CASP)*, Pune, 2016, pp. 315-320.
14. Fang Sheng, A. Bilgin, P. J. Sementilli and M. W. Marcelling, "Lossy and lossless image compression using reversible integer wavelet transforms," *Proceedings 1998 International Conference on Image Processing. ICIP98 (Cat. No.98CB36269)*, Chicago, IL, 1998, pp. 876-880 vol.3.
15. C. Lan, J. Xu, Wenjun Zeng and F. Wu, "Compound image compression using lossless and lossy LZMA in HEVC," *2015 IEEE International Conference on Multimedia and Expo (ICME)*, Turin, 2015, pp. 1-6.
16. G. K. Wallace, "The JPEG still picture compression standard," in *IEEE Transactions on Consumer Electronics*, vol. 38, no. 1, pp. xviii-xxxiv, Feb 1992.
17. A. Skodras, C. Christopoulos and T. Ebrahimi, "The JPEG 2000 still image compression standard," in *IEEE Signal Processing Magazine*, vol. 18, no. 5, pp. 36-58, Sep 2001.
18. M. Rabbani, R. Joshi, "An overview of the JPEG2000 still image compression standard," in Signal Processing: Image Communication, vol. 17, pp.3-48, 2002.
19. S. Ahadpour, Y. Sadra, "Chaotic trigonometric haar wavelet with focus on image encryption," in arxiv.org, arXiV: 1501.00166, 2015.
20. S. Ahadpour, Y. Sadra and M. Sadegh, "Image Encryption Based On Gradient Haar Wavelet and Rational Order Chaotic Maps," in Annals. Computer Science Series, vol. 14, no. 1, pp. 59-66, Jul 2016.
21. N. Ahmed, T. Natarajan and K. R. Rao, "Discrete Cosine Transform," in *IEEE Transactions on Computers*, vol. C-23, no. 1, pp. 90-93, Jan. 1974.
22. W. B. Pennebaker, J. L. Mitchell, "*JPEG still image data compression standard*," Springer (3rd Ed.), ISBN 978-0-442-01272-4, 1993.
23. C. Christopoulos, A. Skodras and T. Ebrahimi, "The JPEG2000 still image coding system: an overview," in *IEEE Transactions on Consumer Electronics*, vol. 46, no. 4, pp. 1103-1127, Nov 2000.
24. D. Taubman, "High performance scalable image compression with EBCOT," in *IEEE Transactions on Image Processing*, vol. 9, no. 7, pp. 1158-1170, Jul 2000.
25. S. G. Mallat, "A theory for multiresolution signal decomposition: the wavelet representation," in *IEEE Transactions on Pattern Analysis and Machine Intelligence*, vol. 11, no. 7, pp. 674-693, Jul 1989.
26. R. C. Gonzalez and R. E. Woods, "Digital Image Processing," in John Wiley & Sons (Second Ed.), Prentice Hall, New Jersey, ISBN 0-201-18075-8 2002.
27. T. Acharya, A. K. Ray, "Image Processing Principles and Applications," John Wiley & Sons, New Jersey, ISBN-13: 978-0471719984, 2005.

28. A. Boggess, F. J. Narcowich, "A First Course in Wavelets with Fourier Analysis," John Wiley & Sons, (2nd Ed.), ISBN: 978-0-470-43117-7, 2009.
29. Z. Wang, E. P. Simoncelli and A. C. Bovik, "Multiscale structural similarity for image quality assessment," *The Thrity-Seventh Asilomar Conference on Signals, Systems & Computers, 2003*, 2003, pp. 1398-1402 Vol.2.
30. Zhou Wang, A. C. Bovik, H. R. Sheikh and E. P. Simoncelli, "Image quality assessment: from error visibility to structural similarity," in *IEEE Transactions on Image Processing*, vol. 13, no. 4, pp. 600-612, April 2004.
31. R. J. Hyndman, A. B. Koehler, "Another look at measures of forecast accuracy," in International Journal of Forecasting, vol. 22, no. 4, Oct 2006.
32. Q. Huynh-Thu and M. Ghanbari, "Scope of validity of PSNR in image/video quality assessment," in *Electronics Letters*, vol. 44, no. 13, pp. 800-801, June 19 2008.